\documentclass[a4paper,conference]{IEEEtran}

\usepackage{amsmath,amssymb,amscd,bbm,amsthm,mathrsfs,dsfont}
\usepackage{algorithmic}
\usepackage{mdwmath}
\usepackage{mdwtab}
\usepackage{bm}
\usepackage{cite}
\usepackage{graphicx,psfrag}
\graphicspath{{Figures/}}
\IEEEoverridecommandlockouts

\usepackage{array}

\usepackage{booktabs}
\usepackage{indentfirst}
\usepackage{amsmath}

\begin{document}

\title{Improving Energy Efficiency in Femtocell Networks: A Hierarchical Reinforcement Learning Framework}

\author{\IEEEauthorblockN{Xianfu Chen$^\ast$, Honggang Zhang$^{\dagger\ddagger}$, Tao Chen$^\ast$, and Mika Lasanen$^\ast$}

\IEEEauthorblockA{
$^\ast$VTT Technical Research Centre of Finland, P. O. Box 1100, FI-90571 Oulu, Finland}

\IEEEauthorblockA{$^\dagger$York-Zhejiang Lab for Cognitive Radio and Green Communications\\
$^\ddagger$Department of ISEE, Zhejiang University, Zheda Road 38, Hangzhou 310027, China\\
Email: xianfu.chen@vtt.fi, honggangzhang@zju.edu.cn, tao.chen@vtt.fi, mika.lasanen@vtt.fi}
}

\maketitle

\begin{abstract}

This paper investigates energy efficiency for two-tier femtocell networks through combining game theory and stochastic learning. With the Stackelberg game formulation, a hierarchical reinforcement learning framework is applied to study the joint average utility maximization of macrocells and femtocells subject to the minimum signal-to-interference-plus-noise-ratio requirements. The macrocells behave as the leaders and the femtocells are followers during the learning procedure. At each time step, the leaders commit to dynamic strategies based on the best responses of the followers, while the followers compete against each other with no further information but the leaders' strategy information. In this paper, we propose two learning algorithms to schedule each cell's stochastic power levels, leading by the macrocells. Numerical experiments are presented to validate the proposed studies and show that the two learning algorithms substantially improve the energy efficiency of the femtocell networks.

\end{abstract}

\section{Introduction}

The rapidly rising energy consumption and greater awareness of environmental protection have created an urgent need for energy-efficient wireless communication techniques. In wireless cellular networks, the radio access part is the main source of energy consumption, accounting for up to more than $70\%$ \cite{Tao}. Therefore, it's crucial to increase the energy efficiency in wireless access networks to meet the challenges raised by the exponential growth in mobile services and energy consumption.

Recently, femtocells have been considered as a promising technique for improving the indoor experience of the mobile users. Due to the short transmit-receive distance property, femtocell technique can greatly lower transmit power, prolong handset battery life, and achieve improved reception \cite{Chandrasekhar}. Although femtocells bring significant benefits to both mobile operators and costumers, to make the deployment feasible, there still exist many technical challenges, e.g., cross-tier/co-tier interference management. In \cite{Vikram}, the authors proposed a distributed utility-based signal-to-interference-plus-noise ratio (SINR) adaptation algorithm to alleviate the cross-tier interference at the macrocell from co-channel femtocells. In \cite{RZhang}, a Stackelberg game was formulated to study the resource allocation in two-tier femtocell networks, where the macro base station (MBS) protects itself by pricing the interference from femtocell users (FUs).

This paper discusses the energy efficiency in femtocell networks. In \cite{Renchao}, the problem of energy-efficient spectrum sharing and power allocation in cognitive radio femtocells was studied, where a three-stage Stackelberg game model was formulated to improve the energy efficiency. In \cite{Ashraf}, the authors proposed a novel energy saving procedure for the femtocell base station (FBS) to decide when to switch on/off. Moreover, we focus mainly on discussing the co-channel operation of femtocells with closed access. The interference in this scenario can not be handled by means of centralized network scheduling, because the number and locations of femtocells are unknown. For such considered networking environment, the femtocells are most likely to be autonomous. Existing research works on reinforcement learning (RL) \cite{Richard} have been carried out in femtocelss networks. In \cite{Giupponi}, a realtime multi-agent RL algorithm that optimizes the network performance by managing the interference in femtocell networks was proposed. In \cite{Bennis}, a distributed learning scheme based on $Q$-learning was developed to manage the femto-to-macrocell cross-tier interference in femtocell networks.

In this paper, we model the energy efficiency aspect of power allocation problem in femtocell networks as a Stackelberg learning game, or leader-follower learning process, with the following characteristics: 1) the macrocells are considered to be the leaders, whereas the femtocells are considered to be the followers; 2) the leaders behave by knowing the response of the femtocells to their own strategy decisions; 3) given the leaders' decisions, the followers compete with each other. Learning is accomplished by directly interacting with the surrounding environment and properly adjusting the strategies according to the realizations of achieved performance. The solution of such a learning game is the Stackelberg equilibrium (SE). If no hierarchy exists during the learning procedure, the Stackelberg learning game reduces to the non-cooperative learning game, which is the scenario discussed in \cite{Qian}.

The rest of this paper is organized as follows. The next section presents the energy efficiency problem in femtocell networks. Section III defines a Stackelberg game theoretic solution for the users' hierarchical behaviors. In Section IV, a Stackelberg learning framework is proposed, and two RL based algorithms are further derived. The numerical results are included in Section V, verifying the validity and efficiency of the proposed algorithms. Finally, we present in Section VI a conclusion of this paper.

\section{Problem Formulation}

\begin{figure}
  \centering
  \includegraphics[width=15pc]{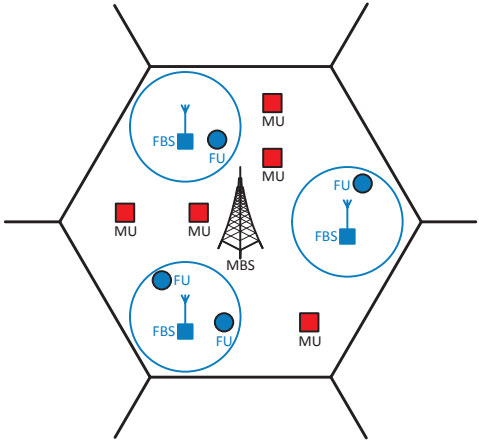}
  \caption{Macrocell is underlaid with femtocells located in the coverage of the macrocell (MBS: macro base station; MU: macrocell user; FBS: femto base station; FU: femtocell user.).}
  \label{pict1}
\end{figure}

The network scenario considered in this paper is shown in Fig. \ref{pict1}, where there exist multiple femtocells and macrocells. Each macrocell consisting of a MBS and multiple macro users (MUs), is underlaid with several co-channel femtocells. In each femtocell, there is one FBS providing service to femtocell users (FUs). Here we assume the closed policy, since customers may prefer that policy because of privacy concerns and limited backhaul bandwidth. Assume same distribution of femtocells in each macrocell, we concentrate on the case of one macrocell for simplification without loss of generality. Suppose $N$ femtocells $B_i(i\geq1)$ operate within the coverage of the macrocell $B_0$. Users of the same macrocell/femtocell adopt time division multiple access (TDMA) for data transmission, thus causing no interference for each other. In the following, this paper mainly addresses the uplink transmissions in the cell over a common spectrum band.

Let $i\in\mathcal{N}$ denote the scheduled user connected to its BS $B_i$, where $\mathcal{N}=\{0,1,\ldots,N\}$ refers to the user index set. Designate the power level of user $i$ as $p_i\in P_i\stackrel{\tiny\mbox{def}}{=}[p_i^{\min},p_i^{\max}]$, the SINR $\gamma_i$ of user $i$ at $B_i$ is given by
\begin{equation*}
    \gamma_i\left(p_i,\textbf{p}_{-i}\right)=\dfrac{h_{i,i}p_i}{\sum_{j\in\mathcal{N}\setminus\{i\}}h_{j,i}p_j+\sigma^2}.
\end{equation*}
Here $-i$ denote all the users in the set $\mathcal{N}$ except user $i$, $\sigma^2$ is the variance of background Additive White Gaussian Noise (AWGN), and $\left\{h_{j,i}\right\}$ is the set of channel gains from user $j$ to $B_i$. Each user aims to improve the energy efficiency, which can be defined as
\begin{equation*}\label{EnergyEfficiency}
    \xi_i\left(p_i,\textbf{p}_{-i}\right)=\frac{W\log_2\left(1+\gamma_i\left(p_i,\textbf{p}_{-i}\right)\right)}{p_i^a+p_i},
\end{equation*}
where $W$ is the spectrum bandwidth, and $p_i^a$ denotes the additional circuit power consumption of user $i$ during transmissions and is independent from the transmission power.

Given the minimum SINR requirement $\gamma_i^*$ chosen to ensure the adequate QoS, we can express the utility function of the user $i$ formally as
\begin{equation*}\label{Utility}
    u_i\left(p_i,\textbf{p}_{-i}\right)=\left\{
    \begin{array}{l@{\quad}l}
        \xi_i\left(p_i,\textbf{p}_{-i}\right),&\mbox{if }\gamma_i\left(p_i,\textbf{p}_{-i}\right)\geq\gamma_i^*;\\
        0, &\mbox{otherwise.}
    \end{array}
    \right.
\end{equation*}
Whenever the macrocell's $\gamma_0^*$ becomes infeasible, MU transmits with maximum power level. We propose that the MBS requests the FUs to reduce their minimum SINR targets. Each user $i\in\mathcal{N}$ tries to find the optimal power level $p_i$ that maximizes its utility,
\begin{equation}\label{Op1}
    \max\limits_{p_i\in P_i}u_i\left(p_i,\textbf{p}_{-i}\right).
\end{equation}

\section{Stackelberg Game Theoretic Solution}

Within the Stakelberg game framework, the MU is modeled as the leader, while the FUs as the followers competing with each other over the spectrum. In order to investigate the existence of an SE, we first define $p_i^*$ to be the best response to $\textbf{p}_{-i}$ if
\begin{align*}
    u_i\left(p_i^*,\textbf{p}_{-i}\right)\geq u_i\left(p_i,\textbf{p}_{-i}\right),\forall p_i\in P_i.
\end{align*}
User $i$'s best response to $\textbf{p}_{-i}$ is denoted as $\mbox{\emph{BR}}_i(\textbf{p}_{-i})$. Let $\mbox{\emph{NE}}(p_0)$ be the Nash equilibrium (NE) strategy of the FUs if the MU chooses to play $p_0$, i.e.
\begin{align*}
    \mbox{\emph{NE}}(p_0)=\textbf{p}_{-0},\mbox{ if }p_i=\mbox{\emph{BR}}_i(\textbf{p}_{-i}),\forall i\in\mathcal{N}\backslash\{0\}.
\end{align*}
The strategy profile $\left(p_0^*,\mbox{\emph{NE}}(p_0^*)\right)$ is an SE if and only if
\begin{align*}
    u_0\left(p_0^*,\mbox{\emph{NE}}(p_0^*)\right)\geq u_0\left(p_0,\mbox{\emph{NE}}(p_0)\right), \forall p_0\in P_0.
\end{align*}

The following theorem establishes the existence of the SE.

\textbf{Theorem 1.} The SE always exists in our defined Stackelberg game with the MU leading and the FUs following.

\emph{Proof}: For $\forall p_0\in P_0$, there is at least one NE in the non-cooperative game $G=\big\langle p_0,\mathcal{N}\backslash\{0\},\{P_i\},\{u_i\}\big\rangle$, since for $\forall i\in\mathcal{N}\backslash\{0\}$
\begin{enumerate}
  \item the strategy profile $\mathcal{P}_i$ is a non-empty, convex, and compact subset of some Euclidean space $\mathfrak{R}^n$;
  \item $u_i$ is continuous in $\textbf{p}_{-i}$ and quasi-concave in $p_i$.
\end{enumerate}

Therefore, there exists $p_0^*\in P_0$, such that
\begin{equation*}
    u_0\left(p_0^*,\mbox{\emph{NE}}\left(p_0^*\right)\right)=\sup\limits_{p_0\in P_0}u_0\left(p_0,\mbox{\emph{NE}}(p_0)\right).
\end{equation*}
We can conclude that the SE always exists in the defined Stackelberg game.$\hfill\blacksquare$

Note that there is only one leader in the Stackelberg game. The MU regards itself as the leader and performs the Stackelberg strategy, and the FUs will act their best responses until reach the equilibrium.

\section{Stackelberg Learning Framework}

In the Stackelberg learning game, each user in the network behaves as an intelligent agent, whose objective is to maximize its expected utility. The game is played repeatedly to achieve the optimal strategies. The Stackelberg learning framework has two levels of hierarchy: 1) the MU aims to maximize its expected utility by knowing the responses of all FUs for each possible play; 2) given the strategy of the MU, the FUs play a non-cooperative game among each other. In this section, we focus our emphasis on how to reach the optimal communication configuration through RL approach \cite{Richard}.

A strategy for user $i\in\mathcal{N}$ at time step $t$ is defined to be a probability vector $\textbf{y}_i^t=\left(y_{i,1}^t,\ldots,y_{i,m_i}^t\right)\in\mathcal{Y}_i$, where $y_{i,j}^t$ means the probability with which the user $i$ chooses action (transmission power) $p_{i,j}\in\mathcal{P}_i$, and $\mathcal{Y}_i$ is the strategy set available to user $i$. Since each user can only choose a power level from a finite discrete set, $\mathcal{P}_i$ is assumed to be a finite set with dimension $m_i$. Then the expected utility for user $i$ at time slot $t$ can be expressed as follows
\begin{align*}
    U_i\left(\textbf{y}_i^t,\textbf{y}_{-i}^t\right)&=\mbox{E}\left[u_i|\mbox{user }j\mbox{ plays strategy }\textbf{y}_j^t,j\in\mathcal{N}\right]\nonumber\\
    &=\sum_{\textbf{p}^t\in\mathcal{P}}u_i\left(\textbf{p}^t\right)\prod_{s\in\mathcal{N}}y_{s,j_s}^t,
\end{align*}
where $\textbf{p}^t=\left(p_{0,j_0}^t,\ldots,p_{N,j_N}^t\right)$ is the vector of actions chosen by all users at time $t$, and $\mathcal{P}=\times_{i\in\mathcal{N}} \mathcal{P}_i$ is the set of all action vectors.

For any stationary strategy\footnote{A strategy is said to be stationary, where $\textbf{y}_i=\left(y_{i,1},\ldots,y_{i,m_i}\right)$ is not changing with time during the stochastic learning process.} of the MU $\textbf{y}_0\in\mathcal{Y}_0$, the best-response strategies of all FUs define an NE strategy $\mbox{\emph{NE}}(\textbf{y}_0)$, i.e. $\mbox{\emph{NE}}(\textbf{y}_0)=\textbf{y}_{-0}^*$, if
\begin{align*}
    \textbf{y}_i^*=\arg\max_{\textbf{y}_i\in\mathcal{Y}_i}U_i\left(\textbf{y}_i,\textbf{y}_{-i}\right),\forall i\in\mathcal{N}\backslash\{0\}.
\end{align*}
The MU's optimal strategy is then
\begin{align*}
    \textbf{y}_0^*=\arg\max_{\textbf{y}_0\in\mathcal{Y}_0}U_0\left(\textbf{y}_0,\mbox{\emph{NE}}(\textbf{y}_0)\right).
\end{align*}
Together $(\textbf{y}_0^*,\mbox{\emph{NE}}(\textbf{y}_0^*))$ constitute a stationary strategy of SE for the Stackelberg learning formulation.

\textbf{Theorem 2.} For the considered Stackelberg learning game, there exist a MU's stationary strategy and a FUs' NE strategy that form an SE.

Inspired by \cite{Yevgeniy}, we can prove Theorem 2 as follows.

\emph{Proof:} If the MU follows a stationary strategy $\textbf{y}_0$, then the Stackelberg learning game is simplified to be a $N$-player stochastic learning game. It has been shown in \cite{Game} that every finite strategic-form game has a mixed strategy equilibrium, that is, there always exists a $\mbox{\emph{NE}}(\textbf{y}_0)$ in our formulation of the discrete power allocation game given the MU's strategy $\textbf{y}_0$. The rest of the proof follows directly from the definition of SE, and is thus omitted for brevity. $\hfill\blacksquare$

\subsection{Reinforcement Learning based Algorithm-I (RLA-I)}
During the Stackelberg learning process, the MU behaves as the leader and knows the strategy information of all FUs. Users with RL ability learn to maximize its individual expected utility through repeated interactions with the networking environment. Among many different implementation of above adaptation mechanism, here we consider the so-called $Q$-learning \cite{Q-learning}, where the users' strategies are parameterized through $Q$-functions that characterize relative expected utility of a particular power level. More specifically, let $Q_i^t\left(p_{i,j_i}\right)$ denote the $Q$-value of user $i$'s corresponding power level $p_{i,j_i}$ at time $t$. Then, after selecting power level $p_{i,j_i}$ at time $t+1$, the $Q$-value is updated according to
\begin{equation}\label{Q-Value-Update}
    Q_i^{t+1}\left(p_{i,j_i}\right)=Q_i^t\left(p_{i,j_i}\right)+\alpha\left(U_i\left(p_{i,j_i},\textbf{y}_{-i}^{t+1}\right)-Q_i^t\left(p_{i,j_i}\right)\right),
\end{equation}
where $\alpha\in[0,1)$ is the learning rate, and
\begin{align}\label{Action-Expected-Utility}
    U_i\left(p_{i,j_i},\textbf{y}_{-i}^{t+1}\right)=\sum_{\textbf{p}_{-i}^{t+1}\in\mathcal{P}_{-i}}u_i\left(p_{i,j_i},\textbf{p}_{-i}^{t+1}\right)\prod_{s\in\mathcal{N} \backslash i}y_{s,j_s}^{t+1}.
\end{align}

Each FU $i\in\mathcal{N}\backslash\{0\}$ doesn't know other competing FUs' private strategies $\textbf{y}_{-(0,i)}^{t+1}$, and the only information it has is the MU's transmission parameters. But the FU $i$ is able to compute the attainable utility $u_i(p_{i,j_i},\textbf{p}_{-i}^{t+1})$ with the feedback information from its receiver. Therefore, FU $i$ can estimate $U_i(p_{i,j_i},\textbf{y}_{-i}^{t+1})$ using
\begin{align}\label{Estimated-Action-Expected-Utility}
    \widetilde{U}_i^{t+1}\left(p_{i,j_i}\right)=\sum_{p_{0,j_0}^{t+1}\in\mathcal{P}_0}y_{0,j_0}^{t+1}\widehat{U}_i^{t+1}\left(p_{i,j_i},p_{0,j_0}^{t+1}\right),
\end{align}
where $\widehat{U}_i^{t+1}\left(p_{i,j_i},p_{0,j_0}^{t+1}\right)$ is given in Eq. (\ref{Estimated-Utility}). $n_{p_{i,j_i},p_{0,j_0}}^t$ is the number of times when MU selects power level $p_{0,j_0}$ and FU $i$ selects power level $p_{i,j_i}$ until time slot $t$.
\begin{figure*}[!t]
\begin{align}\label{Estimated-Utility}
    \widehat{U}_i^{t+1}\left(p_{i,j_i},p_{0,j_0}^{t+1}\right)=\left\{
    \begin{array}{l@{~}l}
        \dfrac{u_i\big(p_{i,j_i},p_{0,j_0},\textbf{p}_{-(i,0)}^{t+1}\big)-\widehat{U}_i^t\left(p_{i,j_i},p_{0,j_0}\right)}
            {n_{p_{i,j_i},p_{0,j_0}}^t+1}+\widehat{U}_i^t\left(p_{i,j_i},p_{0,j_0}\right),&\mbox{if }p_{0,j_0}^{t+1}=p_{0,j_0};\\
        \widehat{U}_i^t\left(p_{i,j_i},p_{0,j_0}^{t+1}\right),&\mbox{otherwise.}
    \end{array}
    \right.
\end{align}
\hrule
\end{figure*}
The updating rule in Eq. (\ref{Q-Value-Update}) can then be rewritten as
\begin{align}\label{Estimated-Q-Value-Update}
    &Q_i^{t+1}\left(p_{i,j_i}\right)=\nonumber\\&~~\left\{
    \begin{array}{l@{~}l}
      Q_i^t\left(p_{i,j_i}\right)+\alpha\left(U_i\left(p_{i,j_i},\textbf{y}_{-i}^{t+1}\right)-Q_i^t\left(p_{i,j_i}\right)\right), &\mbox{if }i=0;\\
      Q_i^t\left(p_{i,j_i}\right)+\alpha\left(\widetilde{U}_i^{t+1}\left(p_{i,j_i}\right)-Q_i^t\left(p_{i,j_i}\right)\right), &\mbox{otherwise}.
    \end{array}
    \right.
\end{align}

Then, we need to specify how the user selects power levels. Greedy selection, when the action with the highest $Q$-value is selected, might generally lead to globally suboptimal solution. Thus, we need to incorporate some way of exploring less-optimal strategies. Here, we focus on Boltamann action selection mechanism, that is, the probability of choosing power level $p_{i,j_i}$ at time $t$ is given by
\begin{equation}\label{Boltzmann}
    y_{i,j_i}^{t+1}=\frac{\exp\left(Q_i^t\left(p_{i,j_i}\right)/\tau_i\right)}{\sum_{l=1}^{m_i}\exp\left(Q_i^t(p_{i,l})/\tau_i\right)},
\end{equation}
where the temperature $\tau_i$ controls the exploration/exploitation tradeoff\cite{Richard}.

Now we present the reinforcement learning based algorithm-I for the Stackelberg learning game.\\
\rule{21pc}{0.1em}
\noindent\emph{RLA-I}

\noindent\rule{21pc}{0.06em}

\vspace{0.1cm}
\noindent\textbf{Initialization:}
\begin{quote}
$t=0$, initialize $Q$-values $Q_i^t(p_{i,j_i})$, for $\forall i\in\mathcal{N}$ and $\forall p_{i,j_i}\in\mathcal{P}_i$.
\end{quote}

\noindent\textbf{Learning:}
\begin{enumerate}
        \item The MU chooses action $p_{0,j_0}$ according to $y_0^t$ and broadcasts its strategy information to all the FUs in the network.
        \item Each FU $i\in\mathcal{N}\backslash\{0\}$ selects action $p_{i,j_i}$ according to $y_i^t$ and sends its relevant strategy information to the MBS.
        \item Users measure their SINR $\gamma_i$ with the feedback information of the intended receiver. If $\gamma_i\geq\gamma_i^*$, then $\xi_i\left(\textbf{p}\right)$ can be achieved; otherwise, the receiver can not receive correctly, thus obtains zero.
        \item The MU updates $U_i\left(p_{i,j_i},\textbf{y}_{-i}^t\right)$ according to Eq. (\ref{Action-Expected-Utility}); and the FUs update $\widetilde{U}_i^t\left(p_{i,j_i}\right)$ basing on Eq. (\ref{Estimated-Action-Expected-Utility}) and Eq. (\ref{Estimated-Utility}).
        \item Update $Q$-values $Q_i^t\left(p_{i,j_i}\right)$ according to Eq. (\ref{Estimated-Q-Value-Update}).
        \item Update the strategies $y_{i,j_i}^{t+1}$ according to Eq. (\ref{Boltzmann}).
        \item $t=t+1$.
\end{enumerate}

\noindent\textbf{End Learning}\\
\noindent\rule[0.1mm]{21pc}{0.1em}

The learning algorithm results in a stochastic process of choosing power level, so we need to investigate the long-term behavior of the learning procedure. Along with the discussion in \cite{Kianercy}, we obtain the following differential equation describing the evolution of the $Q$-values:
\begin{align}\label{Differential-Estimated-Q-Value-Update}
    \frac{\mbox{d}Q_i^t\left(p_{i,j_i}\right)}{\mbox{d}t}=\left\{
    \begin{array}{l@{~}l}
      \alpha\left(U_i\left(p_{i,j_i},\textbf{y}_{-i}^t\right)-Q_i^t\left(p_{i,j_i}\right)\right), &\mbox{if }i=0;\\
      \alpha\left(\widetilde{U}_i^t\left(p_{i,j_i}\right)-Q_i^t\left(p_{i,j_i}\right)\right), &\mbox{otherwise}.
    \end{array}
    \right.
\end{align}
Next, we would like to express the dynamics in terms of strategies rather than the $Q$-values. Toward this end, we differentiate Eq. (\ref{Boltzmann}) with respect to $t$ and use Eq. (\ref{Differential-Estimated-Q-Value-Update}). The set of equations expressed by Eq. (\ref{Differential-Strategy}) can be arrived.
\begin{figure*}
\begin{align}\label{Differential-Strategy}
    \frac{\mbox{d}y_{i,j_i}^t}{\mbox{d}t}=\left\{
    \begin{array}{l@{~}l}
      \dfrac{\alpha}{\tau_0}y_{0,j_0}^t\left\{\left[U_0\left(p_{0,j_0},\textbf{y}_{-0}^{t-1}\right)-\sum\limits_{l=1}^{m_0}y_{0,l}^t
            U_0\left(p_{0,l},\textbf{y}_{-0}^{t-1}\right)\right]-\tau_0\sum\limits_{l=1}^{m_0}y_{0,l}^t\ln\left(y_{0,j_0}^t/y_{0,l}^t\right)\right\}, &\mbox{if }i=0;\\
      \dfrac{\alpha}{\tau_i}y_{i,j_i}^t\left\{\left[\widetilde{U}_i^{t-1}\left(p_{i,j_i}\right)-\sum\limits_{l=1}^{m_i}y_{i,l}^t
            \widetilde{U}_i^{t-1}\left(p_{i,l}\right)\right]-\tau_i\sum\limits_{l=1}^{m_i}y_{i,l}^t\ln\left(y_{i,j_i}^t/y_{i,l}^t\right)\right\}, &\mbox{otherwise}.
    \end{array}
    \right.
\end{align}
\hrule
\end{figure*}

The first term in braces of Eq. (\ref{Differential-Strategy}) asserts that the probability of choosing power level $p_{i,j_i}$ increases with a rate proportional to the overall efficiency of that strategy, while the second term describes the BS's tendency to randomize over possible power levels. Let $Y^t=\left(\textbf{y}_0^t,\ldots,\textbf{y}_N^t\right)$ the strategy profile of all users at time $t$. In the following analysis, we resort to an ordinary differential equation (ODE) whose solution approximates the convergence of $Y^t$\cite{ODE}. Then, as $\alpha^t\rightarrow0$, we can represent the right-hand side of Eq. (\ref{Differential-Strategy}) by a function $f\left(Y^t\right)$. With any initial $Y^0=Y_0$, $Y^t$ will converges weakly, as $\alpha\rightarrow0$, to $Y^*\stackrel{\tiny\mbox{def}}{=}\left(\textbf{y}_0^*,  \mbox{\emph{NE}}\left(\textbf{y}_0^*\right)\right)$ which is the solution of
\begin{align}\label{ODE}
    \frac{\mbox{d}Y}{\mbox{d}t}=f(Y),Y^0=Y_0.
\end{align}

\textbf{Theorem 3.} \emph{RLA-I} will always discover an SE strategy.

\emph{Proof:} We prove this by contradiction. Suppose that the process generated by Eq. (\ref{Boltzmann}) converges to a non-Stackelberg equilibrium. We know that the equilibrium solution of Eq. (\ref{ODE}) that characterize the long term behavior of \emph{RLA-I} are by definition stationary points. This implies that \emph{RLA-I} will only converge to stationary points. This means that stationary points that are not SEs are stable, which contradicting Theorem 2. $\hfill\blacksquare$

\subsection{Reinforcement Learning based Algorithm-II (RLA-II)}

In this subsection, to promote cooperation among the competing FUs, we further design a simple and intuitive rule each FU has that links the other FUs' strategies to its own current transmission strategy. Such a rule reflects an awareness that there are strategic interactions during the learning procedure. FUs with such beliefs do not correctly perceive how the future strategies of their competitors depend on the past. Thus, we propose a conjecture model concerning the way in which the FUs react to each other\cite{Tidball}.

Each FU $i\in\mathcal{N}\backslash\{0\}$ thinks any change in its current transmission strategy will induce other FUs to make well-defined changes in the corresponding time slot. Specifically, we need to express FU $i$'s expected contention measure $b_{i}^{t+1}\big(\textbf{p}_{-(0,i)}^{t+1}\big)=\prod_ {s\in\mathcal{N}\backslash\{0,i\}} y_{s,j_s}^{t+1}$ in Eq. (\ref{Action-Expected-Utility}) through $\tilde{b}_{i}^{t+1}\big(\textbf{p}_{-(0,i)}^{t+1}\big)$, that is
\begin{equation}\label{Conjecture}
    \tilde{b}_i^{t+1}\big(\textbf{p}_{-(0,i)}^{t+1}\big)=b_i^t\big(\textbf{p}_{-(0,i)}^{t+1}\big)-\delta_i\left(y_{i,j_i}^{t+1}-y_{i,j_i}^t\right),
\end{equation}
where $\delta_i>0$ is the belief factor. In other words, every FU $i$ believes that a change of $y_{i,j_i}^{t+1}-y_{i,j_i}^t$ in its strategy at time $t+1$ will induce a change of $\delta_i\left(y_{i,j_i}^{t+1}-y_{i,j_i}^t\right)$ in the expected contention measure exactly related to the transmission strategies of other FUs. It's worth mentioning here that although FU $i$ may be aware that other FUs are subject to many influences on their strategies, when making its own decision, it is only concerned with other FUs' reactions to itself. That means FU $i$ does not take into account whether or not FU $j\left(j\in\mathcal{N}\backslash\{0,i\}\right)$ might react to changes in transmission strategy made by FU $l(l\in\mathcal{N}\backslash\{0,i,j\})$.

Among different possibilities of capturing the expected contention measure $\tilde{b}_i^{t+1}\big(\textbf{p}_{-(0,i)}^{t+1}\big)$, the linear model represented in Eq. (\ref{Conjecture}) is the simplest form based on which one FU can model the impact of its changes in transmission strategy to other competing FUs. The conjecture model deployed by the FUs are based on the concept of reciprocity, which refers to interaction mechanisms in which the FUs repeatedly interact when choosing the power level. If they realize that their probabilities of interacting with each other in the future is high, they will consider their influence on the strategies of other FUs, which is captured in the conjecture model by the positive parameter $\delta_i$. Otherwise, they will act myopically, which is the same learning process as in previous Section IV-A.

Following the previous analysis, Eq. (\ref{Estimated-Action-Expected-Utility}) thus becomes
\begin{align}\label{Estimated-Action-Expected-Utility2}
    \widetilde{U}_i^{t+1}\left(p_{i,j_i}\right)=\sum_{\textbf{p}_{-i}^{t+1}\in\mathcal{P}_{-i}}y_{0,j_0}^{t+1}u_i\left(p_{i,j_i},\textbf{p}_{-i}^{t+1}\right) \tilde{b}_i^{t+1}\big(\textbf{p}_{-(0,i)}^{t+1}\big),
\end{align}
Therefore, we propose \emph{RLA-II} by replacing $\widetilde{U}_i^{t+1}\left(p_{i,j_i}\right)$ in Eq. (\ref{Estimated-Q-Value-Update}) with Eq. (\ref{Estimated-Action-Expected-Utility2}) to discover the SE strategy. \emph{RLA-II} is similar to \emph{RLA-I} except that the FUs update their $Q$-values based on Eq. (\ref{Estimated-Action-Expected-Utility}) in \emph{RLA-I}.

\section{Numerical Results}

We provide insight into the performance comparison of the both learning algorithms through simulation experiments. We consider multiple FUs coexisting with one MU over a spectrum with bandwidth of $1$MHz. The minimum SINR targets of MU and FUs are set to be $3$dB and $5$dB, respectively. The noise of the measurement is according to a zero-mean Gaussian noise with the power of $\sigma^2=-110$dBm. For simplicity, we suppose that the belief factors $\delta_i$ are all equal to $2$, $\forall i\in\mathcal{N}\backslash\{0\}$, i.e., the FUs have the same conjecture ability, and the additional circuit power consumption is $10$dBm for all users. The femtocells are assumed to be uniformly distributed within a circle centered at the MBS with radius of $500$m. The coverage radius of femtocell is $20$m. The channel gains are generated by a log-normal shadowing path loss model, $h_{i,j}=d_{i,j}^{-n}$, where $d_{i,j}$ is the distance between user $i$ and BS $j$, and $n$ is the path loss exponent. In simulation, $n$ is assumed to be 4. The action set of transmission power levels for all users is $\{20,25,30\}$dBm.

\begin{figure}[!t]
  \centering
  \includegraphics[width=20pc]{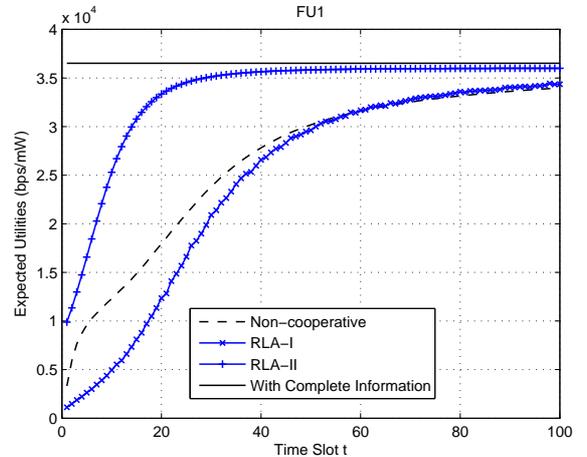}
  \caption{Learning process of the expected utilities for FU 1.}
  \label{pict2}
\end{figure}

\begin{figure}[!t]
  \centering
  \includegraphics[width=20pc]{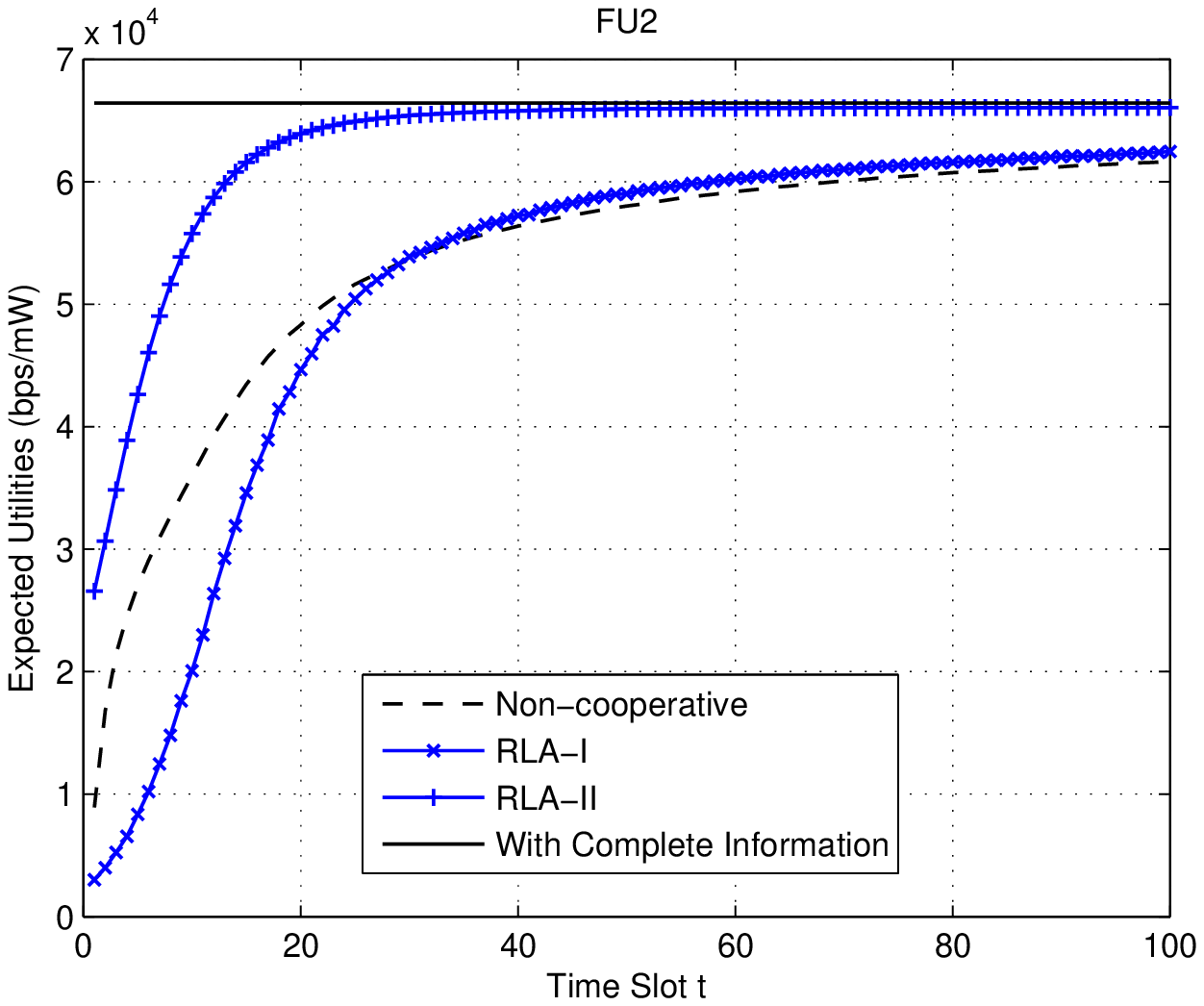}
  \caption{Learning process of the expected utilities for FU 2.}
  \label{pict3}
\end{figure}

\begin{figure}[!t]
  \centering
  \includegraphics[width=20pc]{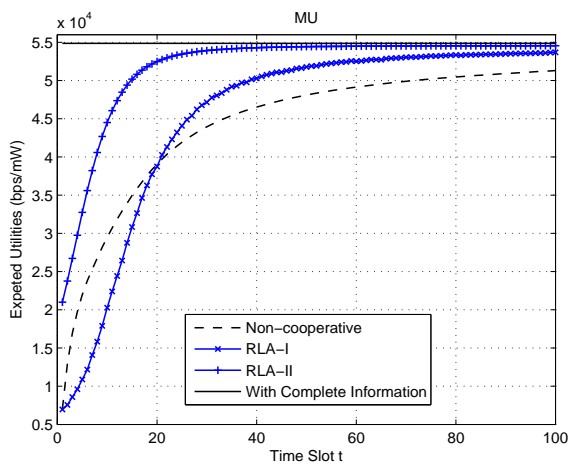}
  \caption{Learning process of the expected utilities for MU.}
  \label{pict4}
\end{figure}

Fig. \ref{pict2}, Fig. \ref{pict3} and Fig. \ref{pict4} show the learning process of expected utilities for each user in the network. The results are compared with
\begin{enumerate}
  \item The fully cooperative power allocation game with complete information case: each user knows all the utility functions and transmission power levels in the network, and then the optimal utilities in the power allocation process can be calculated by each user according to Eq. (\ref{Op1}). This scenario is equivalent to the classic power control game in femtocell networks without the pricing schemes from macrocells as shown in \cite{Vikram}.
  \item The non-cooperative learning process of the power control game without any information exchange case: each user's transmission decisions in the learning process are self-incentive with myopic best response correspondence, which is the scenario discussed in \cite{Qian}.
\end{enumerate}
The first observation from our simulation results is that, whenever we generate random initial probability distributions of the power levels, the equilibrium state of the transmission strategies achieved by all users is independent of these initial values. That is, there exists an SE in the Stackelberg learning game, which confirms Theorem 2.

Secondly, we can find from the curves that the expected utilities of all users in the learning process will finally converge (or approach) to the equilibrium point in the complete cooperation case, and these simulation results validate the conclusion of Theorem 3. In addition, both the proposed reinforcement learning schemes outperform the non-cooperative case, this is because for a Stackelberg learning game, knowing more can improve not only the leader's (MU) own utility, but also the utilities of the followers (FUs). Meanwhile, the \emph{RLA-II} can achieve better performance than \emph{RLA-I}, which is due to the fact that all FUs have the incentive to achieve better utilities thus behave reciprocally at some extent.

\begin{figure}[!t]
  \centering
  \includegraphics[width=20pc]{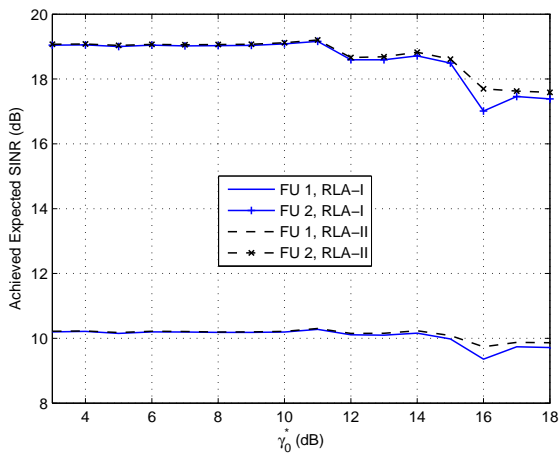}
  \caption{The expected SINRs for FUs versus $\gamma_0^*$.}
  \label{pict5}
\end{figure}

Fig. \ref{pict5} shows the expected SINRs of FUs using \emph{RLA-I} and \emph{RLA-II}, respectively, versus the minimum macrocell QoS requirement $\gamma_0^*$. It is observed that for the same $\gamma_0^*$, the expected SINRs of the FUs with \emph{RLA-II} is in general higher than that with \emph{RLA-I}. This is in accordance with our previous discussions. It is also worth mentioning that when $\gamma_0^*$ is sufficiently large, the expected SINRs of the FUs approach to zero for the two learning algorithms. This is because when $\gamma_0^*$ is sufficiently large, there is no femtocell active in the network.

\section{Conclusion}

In this paper, energy efficiency is investigated for the uplink transmission in a spectrum-sharing-based two-tier femtocell network using stochastic learning theory. The Stackelberg learning framework is adopted to jointly study the utility maximization of the MU and FUs. Based on reinforcement learning, we propose two algorithms, namely, \emph{RLA-I} and \emph{RLA-II}, whose convergence has been further proven theoretically. Numerical experiments illustrate that the reciprocity-inspired \emph{RLA-II} converges more quickly and achieves better performance compared to \emph{RLA-I} and the non-cooperative learning scheme. This comes at the expense of more side information at the FUs. Concludingly, both learning algorithms show the potential in improving the energy efficiency in the femtocell networks.


\begin{thebibliography}{29}

\bibitem{Tao}
T. Chen, Y. Yang, H. Zhang, H. Kim, and K. Horneman, ``Network energy saving technologies for green wireless access networks," \emph{IEEE Wireless Communications}, vol. 18, no. 5, pp. 30-38, October 2011.

\bibitem{Chandrasekhar}
V. Chandrasekhar, J. Andrews, and A. Gatherer, ``Femtocell networks: a survey," \emph{IEEE Communications Magazine}, vol. 46, no. 9, pp. 59-67, September 2008.

\bibitem{Vikram}
V. Chandrasekhar and J. G. Andrews, ``Power control in two-tier femtocell networks," \emph{IEEE Transactions on Wireless Communications}, vol. 8, no. 8, pp. 4316-4328, August 2009.

\bibitem{RZhang}
X. Kang, R. Zhang, and M. Motani, ``Price-based resource allocation for spectrum-sharing femtocell networks: A Stackelberg game approach," \emph{IEEE Journal on Selected Areas in Communications}, vol. 30, no. 3, pp. 538-549, April 2012.

\bibitem{Renchao}
R. Xie, F. R. Yu, and H. Ji, ``Energy-efficient spectrum sharing and power allocation in cognitive radio femtocell networks," in \emph{Proc. of the IEEE International Conference on Computer Communications (INFOCOM)}, Orlando, Florida USA, March 2012.

\bibitem{Ashraf}
I. Ashraf, L. T. W. Ho, and H. Claussen, ``Improving energy efficiency of femtocell base stations via user activity detection," in \emph{Proc. of the IEEE Wireless Communications and Networking Conference (WCNC)}, Sydney, Australia, April 2010.

\bibitem{Richard}
R. S. Sutton and A. G. Barto, \emph{Reinforcement Learning: An Introduction}. Cambridge, MA: MIT Press, 1998.

\bibitem{Giupponi}
L. Giupponi, A. M. Galindo-Serrano, and M. Dohle, ``From cognition to docition: The teaching radio paradigm for distributed \& autonomous deployments," \emph{Computer Communications}, vol. 33, no. 17, pp. 2015-2020, November 2010.

\bibitem{Bennis}
M. Bennis, S. Guruacharya, and D. Niyato, ``Distributed learning strategies for interference mitigation in femtocell networks," in \emph{Proc. of the IEEE Global Communications Conference (GLOBECOM)}, Houston, Texas, USA, December 2011.

\bibitem{Qian}
C. long, Q. Zhang, B. Li, H. Yang, and X. Guan, ``Non-cooperative power control for wireless ad hoc networks with repeated games," \emph{IEEE Journal on Selected Areas in Communications}, vol. 25, no. 6, pp. 1101-1112, August 2007.







\bibitem{Yevgeniy}
Y. Vorobeychik and S. Singh, ``Computing stackelberg equilibria in discounted stochastic games," in \emph{Proc. of the Twenty-Sixth National Conference on Artificial Intelligence}, Toronto, Canada, July 2012.

\bibitem{Game}
D. Fudenberg and J. Tirole, \emph{Game Theory}. Cambridge, MA: MIT Press, 1992.

\bibitem{Q-learning}
C. J. C. H. Watkins and P. Dayan, ``$Q$-learning," \emph{Machine Learning}, vol. 8, pp. 279-292, 1992.

\bibitem{Kianercy}
A. Kianercy and A. Galstyan, ``Dynamics of Boltzmann $Q$-learning in two-player two-action games," \emph{Physical Review E}, vol. 85, no. 4, pp. 041145, April 2012.

\bibitem{ODE}
P. S. Sastry, V. V. Phansalkar, and M. A. L. Thathachar, ``Decentralized learning of Nash equilibria in multi-person stochastic games with incomplete information," \emph{IEEE Transactions on Systems, Man and Cybernetics}, vol. 24 , no. 5, pp. 769-777, May 1994.

\bibitem{Tidball}
A. Jean-Marie and M. Tidball, ``Adapting behaviors through a learning process," \emph{Journal of Economic Behavior \& Organization}, vol. 60, no. 3, pp. 399-422, July 2006.



\end{thebibliography}
\end{document}